\documentclass[10pt,conference]{IEEEtran}
\IEEEoverridecommandlockouts

\usepackage[noadjust]{cite}
\usepackage{amsmath,amssymb,amsfonts}
\usepackage{algorithmic}
\usepackage{graphicx}
\usepackage{textcomp}
\usepackage{xcolor}
\usepackage{url}
\usepackage{booktabs}
\usepackage{tabularx}
\usepackage{multirow}
\usepackage{latexsym}
\usepackage{minted}
\usepackage{indentfirst}
\usepackage{graphicx}
\usepackage{tabularx}
\usepackage{amsmath,amssymb}
\usepackage{booktabs}
\usepackage{array}
\usepackage{multirow}
\usepackage[hidelinks]{hyperref}
\usepackage[table]{xcolor}
\def\BibTeX{{\rm B\kern-.05em{\sc i\kern-.025em b}\kern-.08em
    T\kern-.1667em\lower.7ex\hbox{E}\kern-.125emX}}
\begin{document}

\title{Fake News Theories: Harnessing Disciplinary Insights for Computational Modeling, Detection, and Explanation\\
}

\author{
\IEEEauthorblockN{
Zhaoyang Cao\IEEEauthorrefmark{1},
Miriam Metzger\IEEEauthorrefmark{2},
and Reza Zafarani\IEEEauthorrefmark{1}}
\IEEEauthorblockA{
\IEEEauthorrefmark{1}Electrical Engineering and Computer Science, Syracuse University, Syracuse, NY, USA}
\IEEEauthorblockA{
\IEEEauthorrefmark{2}Department of Communication, University of California, Santa Barbara, Santa Barbara, CA, USA}
\IEEEauthorblockA{
Email: zycao@data.syr.edu, metzger@ucsb.edu, rzafaran@syr.edu}
}

\maketitle

\begin{abstract}
Disinformation research has produced increasingly accurate automated fake-news detectors, but many systems remain difficult to interpret and are weakly connected to established theories of persuasion, credibility, and human judgment. In this paper, we develop a theory-informed computational framework that translates cross-disciplinary theories of fake news into measurable features for automated detection and explanation through statistical techniques and large language models. To that end, we conduct a structured cross-disciplinary review of theories from social sciences, psychology, economics, among other disciplines that reveal how fake news persuades and spreads, thereby establishing a broad theoretical foundation for computational modeling. Experiments on benchmark datasets show that theory-derived features are predictive and provide interpretable, theory-referenced diagnostic signals. Multi-feature models generally outperform individual features, although gains among the strongest small feature combinations are modest. Our work highlights the value of interdisciplinary perspectives in building robust and interpretable fake news detection systems, advancing the foundation for human-centered approaches in combating disinformation.
\end{abstract}


\section{Introduction}
The spread of disinformation and misinformation on digital platforms poses a significant threat to justice, democracy, and public trust (\cite{shu2017fake,zhou2020survey}). Although recent meta-analyses indicate that people can generally discern fake from real news with a fair degree of accuracy \cite{pfander2025spotting}, the massive volume, velocity, and variety of information that circulates online make this task challenging for humans to perform reliably and consistently \cite{demartini2020human}. Moreover, even small errors in detection can have disproportionate societal effects when online disinformation spreads rapidly. As a result, developing automated tools for detecting fake news is essential, not because humans are incapable of judgment, but because manual detection is impractical in today's digital information ecosystem. 

Broadly, existing automated approaches can be grouped into \textit{content-based} and \textit{propagation-based} misinformation detection systems. Recent studies have made significant contributions in both categories (\textit{content-based}: \cite{mikolov2013efficient,pan2018content,kong2020fake,albahar2021hybrid,zhou2020similarity};  and \textit{propagation-based}: \cite{matsumoto2021propagation,meyers2020fake,silva2021propagation2vec,zhou2019network,shu2020hierarchical}). Content-based approaches determine credibility primarily from the information contained in an article including linguistic style, semantics, sentiment, factual consistency, and other textual or multimodal cues. In contrast, propagation-based approaches exploit how information diffuses through a social system, for example through resharing patterns, interaction networks, temporal diffusion structures, or relationships among users and sources. The former can operate directly on static news content, whereas the latter generally requires additional social-network or diffusion metadata. Both paradigms have demonstrated value, but they rely on different forms of evidence. We argue that such techniques should also concentrate on how social theories can help detect fake news with better interpretability. Despite the rapid development of automated fake news detection systems, most approaches are primarily optimized for predictive performance while providing limited interpretability \cite{mishima2022survey}.  Such ``black-box'' models can achieve high accuracy but often fail to provide insights into why a certain piece of news is classified as true or fake. This lack of transparency reduces the practical utility for journalists, policymakers, and fact-checkers, who require interpretable evidence to justify decisions. Furthermore, interpretability is essential for identifying systematic biases in models and ensuring fairness \cite{raj2023true}. In particular, transparent and interpretable systems have great potential to assist domain experts in identifying and intervening in the spread of fake news (\cite{john2025explainability,chien2022xflag}).

To bridge the gap of limited interpretability in existing automated fake news detection systems, recent efforts have begun to incorporate theories from different fields (e.g., communication, psychology, and economics) into fake news detection frameworks. For instance, the work by Zhou \textit{et al.} in fake news early detection proposed leveraging clickbait-related attributes as an early indicator of fake news \cite{zhou2020fake} by drawing upon communication and media psychology theories focused on how online platforms capture and manipulate user attention. To classify clickbait attributes,  relying on social science theories, the researchers applied specific metrics: measuring readability via the Flesch Reading Ease Index and sensationalism through sentiment polarity. Additionally, Qureshi \textit{et al.} proposed a framework that includes theory-driven dimensions such as source trustworthiness and diffusion pattern for the assessment of information credibility. The authors developed a structured framework that can systematically evaluate credibility signals across multiple platforms. By combining computational models with theory, this framework provides an extensible approach to detecting misinformation in fast-paced digital platforms such as Twitter (now X) or Facebook \cite{qureshi2021social}.

The aforementioned studies have demonstrated the utility of theory-motivated psychological, linguistic, and diffusion cues for misinformation assessment. For example, Zhou \textit{et al.} operationalized communication and deception-related theories as computable content attributes, and Qureshi \textit{et al.}  developed a theory-driven credibility framework incorporating source and diffusion dimensions. These studies establish the value of theory-informed modeling, but they consider a more limited set of theoretical perspectives. To overcome these limitations, a more thorough, multi-theoretical perspective is needed to identify the diverse features of deceptive content. 

Fundamental human cognition and behavior theories developed across various social science disciplines provide invaluable insights for fake news analysis. These theories can facilitate building well-justified and explainable models for fake news detection. In our work, we have conducted an extensive cross-disciplinary search to identify well-known theories that have the potential to be used for developing explainable fake news models. We propose a system that leverages representations captured from social science theories and Large Language Models (LLMs) to explore theory-informed signals and evaluate the importance of various features. By integrating insights from multiple disciplinary theories and translating them into measurable computational features through machine learning methods, we aim to improve both detection performance and model interpretability. Accordingly, our overarching research question is: \textit{To what extent can theoretically grounded signals derived from multiple social-science perspectives be transformed into interpretable computational features for distinguishing fake from real news across datasets?} The contributions of this article can thus be summarized as follows: First, we systematically review and integrate theories from social science and psychology that explain why and how fake news persuades or spreads, providing a comprehensive theoretical foundation for computational modeling. Second, we propose a theory-driven feature system that translates conceptual principles into measurable computational features using statistical and machine learning approaches. Last, we empirically validate the theory-informed features within a unified fake news detection model to evaluate their predictive effectiveness and interpret their relationships across different theoretical dimensions \footnote{The code and data are available at \mbox{\url{https://anonymous.4open.science/r/TheoriesOnDisinformation}}}.

\section{Theoretical Foundation}

\begin{table*}[t]
\centering
\caption{Theoretical predictions to help distinguish fake from real news.}
\footnotesize
\renewcommand{\arraystretch}{1.22}
\setlength{\tabcolsep}{4pt}
\begin{tabularx}{0.98\textwidth}{|
    >{\centering\arraybackslash}p{1.4cm}
  | >{\arraybackslash}p{5.4cm}
  | X |}
\hline
\rowcolor{gray!20}
\textbf{Category} & \textbf{Theory/Concept} & \textbf{Compared to real news, fake news should \ldots} \\
\hline

\multirow[c]{9}{*}{\raisebox{-11.8em}{Content}}
  & Information Processing Theory of Language Intensity Effects \cite{hamilton1993extending}; \newline Confidence Heuristi \cite{thomas1995confidence}; \newline Scarcity Principle \cite{mittone2009scarcity}
  & use more intense, sensational, or hyperbolic language and more certain language. \\
\cline{2-3}

  & Emotion Theories \cite{izard2009emotion}; \newline Negativity Bias \cite{rozin2001negativity}
  & contain more emotive language, and especially contain negative emotions. \\
\cline{2-3}

  & Expectancy Violation Heuristic \cite{gass2022persuasion, metzger2010social}
  & contain less clear, organized, and accurate (i.e., error-free) writing. \\
\cline{2-3}

  & Narrative Paradigm \cite{fisher1984narration, metzger2003credibility}
  & be less coherent and have internal logical contradictions. \\
\cline{2-3}

  & Sourcing Theory \cite{metzger2003credibility}
  & include less supporting evidence and attributions to external sources (e.g., citations, quotes, links, data). \\
\cline{2-3}

  & Illusory Truth Effect \cite{hasher1977frequency}
  & have more repetitive content. \\
\cline{2-3}

  & Linguistic Differences \cite{lugea2021linguistic}; \newline Cognitive Load Theory \cite{sarzynska2023truth}
  & possess less lexical and syntactic complexity; more personal pronouns, verbs, adverbs, function words, modal words, negations, but fewer nouns and named entities. \\
\cline{2-3}

  & Information Manipulation Theory -- Maxim of Manner \cite{mccornack1992information}
  & contain more vague or ambiguous expressions. \\
\cline{2-3}

  & Uncertainty Reduction Theory \cite{knobloch2008uncertainty}
  & include more uncertainty-inducing cues in headlines, such as question marks, ellipses, exclamation marks, and surprising topics. \\
\hline

\multirow{1}{*}{\raisebox{-1.2em}{Source}}
  & Source Credibility Theory \cite{hovland1951influence}; \newline Authority Heuristic \cite{wilson1983second}
  & come from less reputable or known sources. \\
\hline

\multirow{1}{*}{\raisebox{-0.7em}{Network}}
  & Consensus Heuristic \cite{metzger2010social}; \newline Homophily Principle \cite{mcpherson2001birds}
  & exhibit less external consistency or consensus across different outlets or sources, and show more linkages to known fake news sites. \\
\hline
\end{tabularx}
\label{table1-theory}
\end{table*}

We start by discussing the cross-disciplinary fundamental theories relevant to fake news detection that we identified in our review of the social and economic sciences literature. These theories can help identify potential differences between real and fake news stories. Undergirding such theories is consideration of the motivations underlying the production and dissemination of real versus fake news. Although financial, informational, and persuasive motivations likely underlie both types of news to some extent, true news places more weight on informing citizens, whereas fake news primarily aims to attract attention and persuade audiences for the purpose of financial or political gain \cite{allcott2017social}. As such, theories concerning attention and persuasion are particularly useful in explaining why and how fake news is likely to differ from real news. Before delving into the details of these theories, we first offer a categorization to help organize our presentation of the theories. 

Social science theories helpful to detecting fake news can be classified as those focusing on (I) the \textit{news content} itself, (II) its \textit{source}, and (III) \textit{network} factors.

\noindent \textbf{I. \textit{Content-based theories.}}  
Theories that focus on the content of messages can help illuminate potential differences between real and fake news stories. For the reasons stated above, fake news is more likely to contain content that garners attention and employs techniques to enhance persuasion than real news. Indeed, several content features have been theorized to distinguish low from high credibility messages and/or to boost persuasion in previous work, for example, language intensity, emotional tone, information quality and objectivity, internal consistency, presence of evidence and attribution, and repetition.

\noindent \textbf{II. \textit{Source-related theories.}}  
Rather than focusing on message content, these theories consider aspects of the source, author, or origin of information that are useful for determining the credibility of a message. \textit{Source credibility theory} is the primary framework employed in this area of research. The theory states that source credibility affects the believability of the source’s message; thus, source credibility is key within the persuasion research literature.

\noindent \textbf{III. \textit{Network-related theories.}} 
Theories that focus on the relationships between information and its sources are also useful for differentiating fake news from real news. The most relevant theories are the \textit{consensus heuristic}, which predicts that agreement between different pieces of information or sources signals truth, and the \textit{homophily principle}, which predicts greater network connections between similar entities \cite{xie2025survey}. The dynamics of information propagation within social networks provide cues for assessing veracity \cite{bian2020rumor}. False information often spreads rapidly with targeted dissemination strategies and between sources with similar dubious credibility. Thus, analyzing these patterns can help detect misinformation. Network-level theories are retained in the theoretical reviews but are \textit{not empirically evaluated} because the benchmark datasets do not provide the propagation traces or inter-source network information.

The structured cross-disciplinary review covered research in communication, psychology, information systems, economics, and computational misinformation research. Candidate theories were identified from prior fake-news and credibility reviews. We retained a theory or theoretical concept when it satisfied the following criteria: (1) it provided an explicit prediction by content, source, information processing, persuasion, or information diffusion; and (2) the factor identified by the theory could differentiate credible from deceptive information. Table~\ref{table1-theory} summarizes several promising social science theories and presents their specific predictions on how fake news differs from real news. Due to space limitations, detailed explanations of each theory are provided in \href{https://docs.google.com/document/d/e/2PACX-1vRMkeNCNq_wgvkoWqxAHcdmDfPObzsxsrLGKopUyRM_-_IozWaJirBe4kzI7sRySZ7lg6Cqzbwc9_od/pub}{\textcolor{blue}{Appendix A}}. In summary, the foregoing theories and research findings suggest several ways in which real news might be differentiated from fake news.

\section{Theory-Informed Computational Framework}
To build a fake news detection system to harness theory for accurate detection, we rely on social science theories that are both (a) applicable and (b) data-testable to distinguish fake from real news. The procedures for building systematic features that can reliably capture these theories are described in detail in Section \ref{3.2-feature}. Our goal was to include a diverse and representative set of theories to capture multiple dimensions of fake news dynamics. Before further explanation, we formulate the problem as follows:

\subsection{Problem formulation}
In our system, each candidate news text is encoded as an $n$-dimensional representation $\mathbf{e}\in\mathbb{R}^n$, where each component of $\mathbf{e}$ corresponds to theory-driven representations (i.e., data features). The detection task is to learn the mapping function $\mathcal{F}$ such that 
\begin{equation}
\mathcal{F}:\mathbf{e} \xrightarrow{TD} \hat{\mathbf{y}}
\end{equation} where $\hat{\mathbf{y}}\in\{0,1\}$ is the predicted news label; $\hat{\mathbf{y}}=1$ denotes a predicted true article, and $\hat{\mathbf{y}}=0$ denotes a fake article. We rely on the $TD$ (training dataset) $TD = \{\mathbf{e}^{(k)},\mathbf{y}^{(k)} | \mathbf{e}^{(k)}\in\mathbb{R}^n,\mathbf{y}^{(k)}\in\{0,1\}, k\in \mathbb{N} \}$ to train the mapping function $\mathcal{F}$, which provides both the theory-informed features representations and the ground-truth label to estimate the parameters.

Within the classical supervised-learning paradigm, both the expressiveness and interpretability of our system depend on (i) how a news representation is constructed and (ii) the design of the classifier $\mathcal{F}$. We elaborate on these two components in Section \ref{3.2-feature}.

\subsection{News Representation and Classification}\label{3.2-feature}

\begin{table*}[ht]
\centering
\caption{Mapping between theory-informed representations and the social-science theories in Table~\ref{table1-theory}.}
\footnotesize
\renewcommand{\arraystretch}{1.18}
\setlength{\tabcolsep}{4pt} 
\begin{tabularx}{0.98\textwidth}{
  >{\raggedright\arraybackslash}p{0.24\textwidth}
  >{\raggedright\arraybackslash}p{0.54\textwidth}
  >{\centering\arraybackslash}p{0.14\textwidth}
  >{\centering\arraybackslash}p{0.06\textwidth}
}
\rowcolor{gray!15}
\toprule
\textbf{Representation} &
\textbf{Corresponding Theory/Concept (Table~\ref{table1-theory})} &
\textbf{Category}\\
\midrule
Confidence Heuristic & Information Processing Theory of Language Intensity; Confidence Heuristic & Content\\

Scarcity Principle & Scarcity Principle & Content\\

Negativity Bias & Emotion Theories; Negativity Bias & Content \\

Coherence & Expectancy Violation Heuristic & Content\\

Logical Consistency & Narrative Paradigm & Content \\

Supporting Evidence & Sourcing Theory& Content  \\

Illusory Truth Repetition & Illusory Truth Effect & Content\\

Information Complexity & Linguistic Differences; Cognitive Load Theory & Content\\

Language Vagueness & Information Manipulation Theory — Maxim of Manner & Content \\

Information Gap & Uncertainty Reduction Theory & Content \\

Source Credibility &  Source Credibility Theory; Authority Heuristic & Source\\

\addlinespace[0.2em]
\rowcolor{gray!08}
— & Consensus Heuristic; Homophily Principle & Network \\
\bottomrule
\end{tabularx}
\label{table2:theory-map}
\end{table*}

Each aforementioned theory suggests important cues in both true and fake news articles, which if captured systematically and efficiently using computational techniques, can help detect fake news. Our goal in this section is to detail the computational techniques we used to extract the theory-driven representations.

To make the link between our features and their theoretical foundations clear, Table \ref{table2:theory-map} provides a mapping from each representation to the social-science theories/concepts summarized in Table \ref{table1-theory}. While Table \ref{table1-theory} presents the broader set of theories identified in our cross-disciplinary review, not all of these theories can be directly operationalized using the information available in benchmark datasets. We then select the theories for empirical evaluation whose key concepts can be transformed into measurable computational representations from news content or available source information. Specifically, these theories motivate ten content-based representations capturing observable textual characteristics, together with a source-related representation when publisher information is available. Table \ref{table2:theory-map} summarizes this theory-to-representation mapping and indicates the theories that are empirically tested in our system. We explicitly focus on content-related and source-related theories, given that our data consist solely of static news articles without user-level interactions or social diffusion records.
\newline\noindent $\blacktriangleright$ To model the \textbf{\textit{confidence heuristic}}, we assign a binary label to indicate whether an article is overall \texttt{`certainty' (1)} or \texttt{`uncertainty' (0)}. Specifically, we employ Linguistic Inquiry and Word Count \cite{pennebaker2001linguistic} to derive associated words of certainty and uncertainty. An article is classified as \texttt{`certain'} when the frequency of certainty-related words is higher than that of uncertainty-related words, and \texttt{`uncertain'} otherwise. 
\newline\noindent $\blacktriangleright$ For the \textbf{\textit{scarcity principle}}, we generate a binary label using LLMs guided by prompt engineering. We use `\texttt{gemma-2-9b-it}', and the prompt is constructed based on the definitions from the original paper that introduced the scarcity principle (see detailed prompt in \href{https://docs.google.com/document/d/e/2PACX-1vRMkeNCNq_wgvkoWqxAHcdmDfPObzsxsrLGKopUyRM_-_IozWaJirBe4kzI7sRySZ7lg6Cqzbwc9_od/pub}{\textcolor{blue}{Appendix C}}).
\newline\noindent $\blacktriangleright$ For \textbf{\textit{negativity bias}}, we employ a valence-aware dictionary and sentiment reasoner (\texttt{VADER}), a widely used tool for sentiment analysis. By analyzing the emotional polarity of single words and aggregating their scores, the compound value of an article is derived and used to measure the overall sentiment. The compound value ranges from -1 to +1, where scores above 0.5 indicate positive sentiment, scores below -0.5 represent negative sentiment, and scores in between are regarded as neutral.
\newline\noindent $\blacktriangleright$ To operationalize textual \textbf{\textit{coherence}}, we adopt the \textit{entity grid model}, which captures how entities are distributed and transition across consecutive sentences. Specifically, for each entity mentioned in the text, we track its syntactic role (e.g., \textit{Subject (S)}, \textit{Object (O)}, or \textit{Other (X)}) in every sentence. This yields a sequence of role transitions, such as $S \rightarrow O \rightarrow S$, representing how the entity’s discourse function evolves over sentences. We then compute the probability of each observed transition pattern based on a predefined distribution of entity-role transitions. To quantify overall coherence of the article, we use the \texttt{average log-probability} of all transitions across entities:
\begin{equation}
\texttt{c} = \frac{1}{N} \sum_{i=1}^{N} \log(p_i),
\end{equation}
where $N$ denotes the number of entity transitions and $p_i$ is the estimated probability of the $i$-th transition. A score closer to zero indicates higher coherence, as most transitions align with dominant discourse patterns.
\newline\noindent $\blacktriangleright$ To model \textbf{\textit{logical consistency}}, we computed a numerical score as the representation of an article. This score is defined as the proportion of consecutive sentences that exhibit logical consistency. We use a pre-trained model (\texttt{roberta-large-mnli}) trained using Multi-Genre Natural Language Inference to obtain this score. The language model will classify adjacent sentence pairs as one of: \texttt{ENTAILMENT}, \texttt{CONTRADICTION}, or \texttt{NEUTRAL}. We only count sentence pairs that show an \texttt{ENTAILMENT} relationship.
\newline\noindent $\blacktriangleright$ For \textbf{\textit{supporting evidence}}, we define a heuristic approach that aggregates multiple in-text evidence signals. Specifically, we operationalize five \textit{sub-features}: (i) the frequency of URLs, (ii) quotations, (iii) temporal references (e.g., years, months, or dates), (iv) attribution phrases (e.g., “according to,” “reported by”), and (v) the proportion of URLs from traditionally trustworthy domains (e.g., \texttt{.gov}, \texttt{.edu}). Each sub-feature is first converted into a density measure (occurrences per thousand words) and then normalized. The final representation is computed as the mean of the normalized sub-feature activations.
\newline\noindent $\blacktriangleright$ To capture the \textbf{\textit{illusory truth effect}}, we measure how frequently claims are repeated within the news article. To extract this feature, we employ `\texttt{gpt-4o-mini}' as a semantic reasoning module. Specifically, given the full article text, we prompt the model to (i) identify distinct claims presented in the news article, and (ii) count the number of times each claim is repeated by sentences that express the same or equivalent meaning (see detailed prompt in \href{https://docs.google.com/document/d/e/2PACX-1vRMkeNCNq_wgvkoWqxAHcdmDfPObzsxsrLGKopUyRM_-_IozWaJirBe4kzI7sRySZ7lg6Cqzbwc9_od/pub}{\textcolor{blue}{Appendix C}}). Following this process, we compute the total number of repetitions across all claims, excluding the initial mention of each claim. A higher score indicates stronger use of repetitive persuasion and is expected to be more prevalent in fake news.
\newline\noindent $\blacktriangleright$ To capture  \textbf{\textit{information complexity}}, we leverage the \textit{Flesch-Kincaid Reading Ease} score, a metric that quantifies textual complexity \cite{flesch1948new}. Higher reading scores indicate greater complexity and weaken the readability of the article.
\newline\noindent $\blacktriangleright$ Annotating \textbf{\textit{language vagueness}} is challenging because vague expressions can manifest at both the lexical and sentential levels. To construct a reliable corpus for vagueness modeling, we adopt the dataset released by Logan Lebanoff and Fei Liu \cite{lebanoff2018automatic}. The corpus comprises privacy policies of 100 websites with an average length of 2.3K words. Human annotators identified vague terms and assigned each sentence a vagueness rating from 1 to 5, where 1 indicates extreme clarity, and 5 indicates extreme vagueness. In our system, we preprocess this vagueness dataset by retaining only the text and vagueness annotations. We take the mean of the discrete scores that each sentence received and aggregate them to generate a continuous-valued vagueness score for each sentence. Then, we fine-tune a BERT-based classifier on this preprocessed dataset to capture the linguistic patterns associated with vagueness. Once trained, this model is applied to our target news articles, and we can obtain the softmax probability of the \textit{language vagueness}, which serves as an input feature in our downstream task.
\newline\noindent $\blacktriangleright$ 
Last, we obtain the \textbf{\textit{information gap (IG)}} feature. This feature consists of two components: (i) \texttt{punctuation-driven} curiosity cues at the end of the headline and (ii) topic-level \texttt{surprisingness} relative to a corpus baseline.
Given a headline $\mathrm{x}$, we compute a numerical embedding as
\begin{equation}\label{equation2}
    \mathrm{IG}(\mathrm{x})
= w_{\text{P}}\cdot \mathrm{P}(\mathrm{x})
+ w_{\text{S}}\cdot \mathrm{S}(\mathrm{x}),
\end{equation}
where $\mathrm{P}(x)$ refers to the punctuation score of the headline $\mathrm{x}$, $\mathrm{S}(x)$ refers to the suprisingness score of the headline $\mathrm{x}$, and $w_{\text{P}}, w_{\text{S}}>0$ are fusion weights for $\mathrm{P}(x)$ and $\mathrm{S}(x)$ respectively (default $w_{\text{P}}{=}w_{\text{S}}{=}0.5$). The final representation $\mathrm{IG}(\mathrm{x})$ can be derived through the following three steps.
\newline\textbf{(1) Punctuation-driven signal.}
Let $\mathrm{tail}(x)\in\{\texttt{?},\texttt{!},\texttt{...},\texttt{.},\varnothing\}$ denote the trailing punctuation of $\mathrm{x}$ that prioritizes `\textbf{...}' over single-character punctuation.
We define the punctuation score as follows:
\begin{equation}
\mathrm{P}(\mathrm{x})=
\begin{cases}
1, & \text{if }\mathrm{tail}(\mathrm{x})\in\{\texttt{?},\texttt{!},\texttt{...}\},\\[2pt]
0, & \text{if }\mathrm{tail}(\mathrm{x})=\texttt{.},\\[2pt]
\sum\limits_{\mathrm{c}\in\{\texttt{?},\texttt{!},\texttt{...}\}} \mathrm{p}(\mathrm{c}\,|\,\mathrm{x}), & \text{if }\mathrm{tail}(\mathrm{x})=\varnothing.
\end{cases}\label{equation3}
\end{equation}
When $\mathrm{tail}(x)=\varnothing$, we employ the LLM (`\texttt{gemma-2-9b-it}' ) to predict and score the next punctuation over a restricted candidate set $\mathcal{C}=\{\texttt{?},\texttt{!},\texttt{...},\texttt{.}\}$. For single-character punctuations $\mathrm{c}\in\{\texttt{?},\texttt{!},\texttt{.}\}$, let $\mathcal{L}_c(\mathrm{x})$ be the next-token logit. For the ellipsis `\textbf{...}', which is a multi-token sequence $(\mathbf{e_1},\mathbf{e_2},\mathbf{e_3})$ under the tokenizer, we use the autoregressive sequence log-probability
\begin{equation}
\mathcal{L}_{\textbf{...}}(\mathrm{x}) = \sum_{\mathrm{t}=1}^{3} \log p\!\left(\mathbf{e_t} \,\middle|\, \mathrm{x}, \mathbf{e_{1:{t-1}}}\right),
\end{equation}
Then a restricted softmax over $\mathcal{C}$ is defined as:
\begin{equation}
\begin{split}
\mathrm{p}(\mathrm{c}\,|\,\mathrm{x})
&=\frac{\exp\big(\tilde{\mathcal{L}}_c(\mathrm{x})\big)}
       {\sum_{\mathrm{c}'\in\mathcal{C}}
        \exp\big(\tilde{\mathcal{L}}_{\mathrm{c}'}(\mathrm{x})\big)},\\[6pt]
\tilde{\mathcal{L}}_\mathrm{c}(\mathrm{x})
&=\begin{cases}
    \mathcal{L}_\mathrm{c}(\mathrm{x}), 
      & \mathrm{c}\in\{\texttt{?},\texttt{!},\texttt{.}\},\\
    \mathcal{L}_{\textbf{...}}(\mathrm{x}), 
      & \mathrm{c}=\texttt{...}
  \end{cases}
\end{split}
\end{equation}
$\mathrm{P}(x)$ is equivalent to the model’s total probability mass assigned to curiosity-inducing endings (\texttt{?}/\texttt{!}/\texttt{...}) as the immediate next punctuation.
\newline\textbf{(2) Topic surprisingness.}
Let a Latent Dirichlet Allocation model generate a topic-mixture distribution $\mathcal{P}(\mathrm{x})$ for headline $\mathrm{x}$. 
We estimate a \emph{baseline} topic distribution $\mathcal{Q}$ by averaging topic-mixture distributions over the training corpus $\{\mathrm{x}_i\}_{i=1}^{N}$:
\begin{equation}
\mathcal{Q} \;=\; \frac{1}{N}\sum_{\mathrm{i}=1}^{N} \mathcal{P}(\mathrm{x}_\mathrm{i})
\end{equation}
The divergence of $\mathcal{P}(\mathrm{x})$ from $\mathcal{Q}$ is measured by the Kullback–Leibler (KL) divergence:
\begin{equation}
\mathcal{D}_{\mathrm{KL}}\!\big(\mathcal{P}(\mathrm{x})\,\|\,\mathcal{Q}\big)
= \sum_{k=1}^{\mathcal{K}} \mathcal{P}_k(\mathrm{x})\,\log\!\frac{\mathcal{P}_k(\mathrm{x})}{\mathcal{Q}_k}
\end{equation}
We transform this non-negative value to $[0,1]$ by a sigmoid variant:
\begin{equation}
\mathrm{S}(\mathrm{x})
= \frac{\mathcal{D}_{\mathrm{KL}}(\mathcal{P}(\mathrm{x})\,\|\,\mathcal{Q})}{1 + \mathcal{D}_{\mathrm{KL}}(\mathcal{P}(\mathrm{x})\,\|\,\mathcal{Q})}.
\end{equation}
\newline\textbf{(3) Fusion Representation.}
The overall \textit{information gap} feature is the linear fusion as defined in Equation \ref{equation2}. A larger value of $\mathrm{S}(\mathrm{x})$ indicates that the content is more surprising, while a higher value of $\mathrm{P}(\mathrm{x})$ indicates greater salience or importance.
\newline\noindent $\blacktriangleright$ Building on source credibility theory and the authority heuristic,  \textbf{\textit{source credibility}} captures the reliability of a news outlet. To encode this dimension, we leverage publisher-level reliability annotations available in each dataset and transform publication venues into numerical identifiers through one-hot encoding. Categorical encoding of publication venues provides a representation capturing latent source-specific credibility. This encoding enables the model to learn associations between publishers and the likelihood of misinformation dissemination.

After detailing the computational extraction of representations derived from social science theories as described above, we fed the theory-driven representations to XGBoost for the purpose of fake news detection. The remaining sections present the results and corresponding analyses.

\section{Evaluation and Results}
We conducted extensive experiments to evaluate our proposed system. The experimental setup is introduced in section \ref{4.1-setup}, and we evaluate our theory-driven system through a series of analyses. Specifically, we test the significance of individual theories, their cross-dataset stability, their relative contributions, and their interactions.

\subsection{Experimental setup}\label{4.1-setup}
For each experiment, we report the average AUC and macro-averaged $F_1$ scores. During text preprocessing, the paragraph separators `$\backslash n$' were removed if they existed, and all text was converted to lowercase to ensure consistency.

Our experiments are conducted on three public benchmark datasets for fake news detection: \textsc{ReCOVery} \cite{zhou2020recovery}, \textsc{Fake\_And\_Real\_News} \cite{mcintire2020fake}, and \textsc{GossipCop} \cite{shu2020fakenewsnet}. The basic statistics of the datasets and detailed dataset descriptions are in Table \ref{table:datasetstat} and \href{https://docs.google.com/document/d/e/2PACX-1vRMkeNCNq_wgvkoWqxAHcdmDfPObzsxsrLGKopUyRM_-_IozWaJirBe4kzI7sRySZ7lg6Cqzbwc9_od/pub}{\textcolor{blue}{Appendix B}}.

\subsection{Baseline Comparison and Predictive Trade-off}
We compare the proposed theory-informed representation with a standard content-based baseline using TF-IDF representations. To ensure a controlled comparison, both the baseline and our theory-informed representation are evaluated using the same XGBoost classifier, training/test partitions, and evaluation metrics.

The TF-IDF baseline achieves an AUC of $0.9553 \pm 0.0046$ and an $F_1$ score of $94.71$. The higher predictive performance of TF-IDF is expected given the different objectives and information available. TF-IDF utilized the complete lexical distribution of an article, potentially capturing a large number of dataset-specific words, topics, entities, and stylistic patterns. In contrast, our theory-informed model compresses each article into a small set of interpretable features, each corresponding to a social-science theory. Our objective is therefore not to outperform a full-text lexical classifier, but to examine how much predictive information can be captured by theoretically motivated and human-interpretable signals. From this perspective, the theory-informed model provides a complementary trade-off: it sacrifices some predictive performance in exchange for a more compact, theoretically grounded, and interpretable representation of the detection process.

\begin{table}[tbp]
\caption{Dataset statistics}
\centering\vspace{-1em}
\setlength{\tabcolsep}{3.8pt}      
\small                              
\begin{tabularx}{\columnwidth}{l|l|>{\centering\arraybackslash}X|>{\centering\arraybackslash}X}
    \hline\hline
    Dataset & Labels & Train & Test \\
    \hline
    \multirow{2}{*}{\textsc{ReCOVery}}        & Truth &  966 & 120 \\
                                               & Fake  &  487 &  64 \\
    \hline
    \multirow{2}{*}{\textsc{Fake\_And\_Real\_News}} & Truth & 1143 & 597 \\
                                               & Fake  & 1154 & 551 \\
    \hline
    \multirow{2}{*}{\textsc{GossipCop}}     & Truth &   6515  & 3286  \\
                                               & Fake  &1653   &798   \\
    \hline\hline
\end{tabularx}
\label{table:datasetstat}\vspace{-1em}
\end{table}

\subsection{Individual importance of theories}
In our experiments, we first evaluated the performance of each theory (that is, its corresponding theory-informed features) individually for fake news detection. Subsequently, we integrated all ten features by concatenation and trained a unified model. The results demonstrate that while individual features typically achieved AUC values between 50.0\% and 70.5\% and $F_1$ scores between 47.9\% and 67.5\%, the integrated model reached an AUC of 79.1\% and an $F_1$ of 76.8\% on the \textsc{ReCOVery} dataset. 
These findings highlight that core social science theories can be computationally tested and effectively utilized for fake news detection. Moreover, in our empirical evaluation, the combination of all theory-driven features not only led to improved predictive performance but also enhanced the interpretability of the model, thereby offering more comprehensive insights into the underlying mechanisms of misinformation propagation.

\begin{figure}[tbp]
  \centering
  \makebox[\columnwidth][c]{%
  \includegraphics[width=1.06\columnwidth]{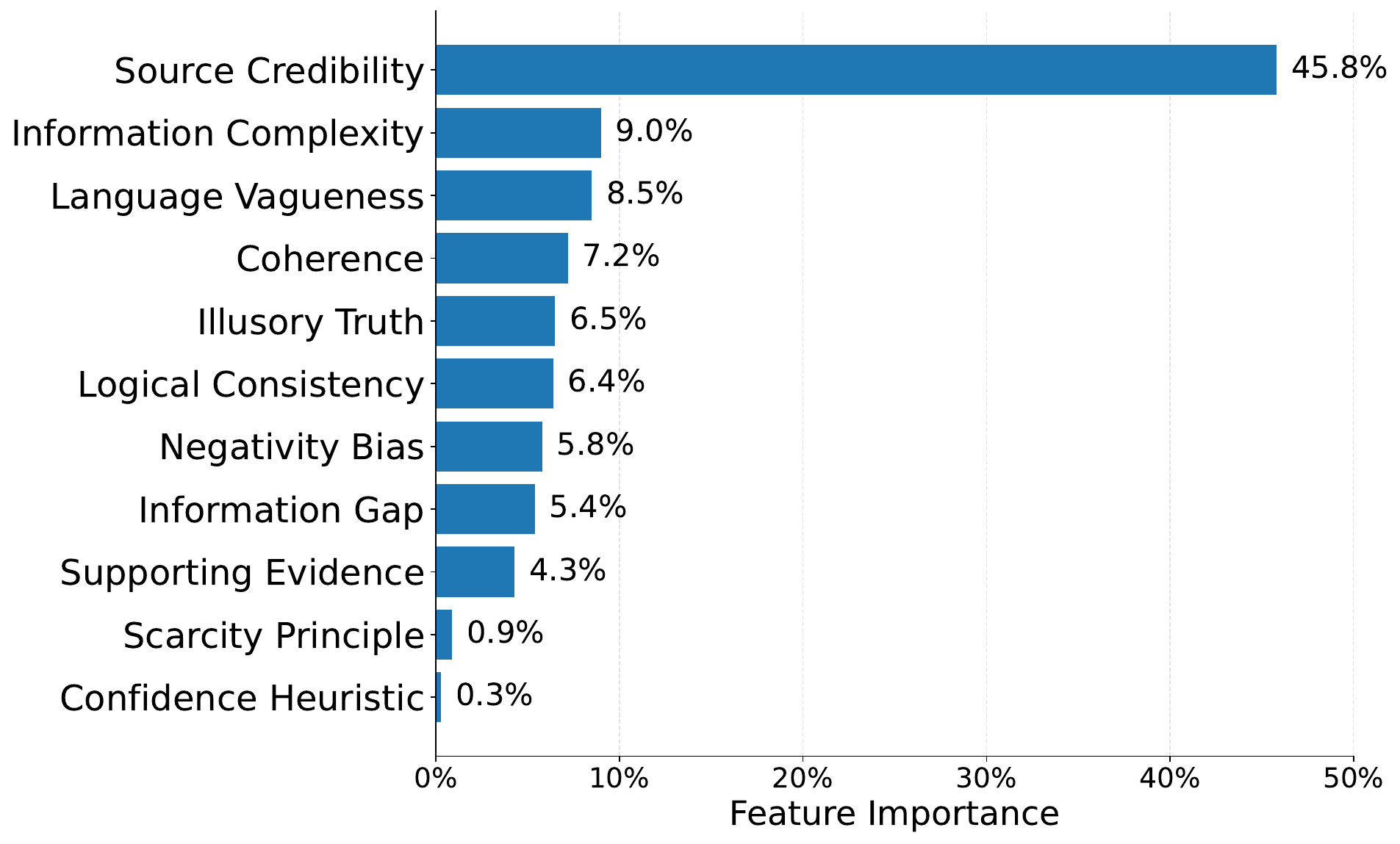}}\vspace{-1em}
  \caption{Feature importance ranking on the \textsc{ReCOVery} dataset. Source Credibility emerges as the most dominant predictor. } 
  \label{rank} 
\end{figure}

\begin{figure}[tbp]
  \centering
  \makebox[\columnwidth][c]{%
  \includegraphics[width=1.02\columnwidth]{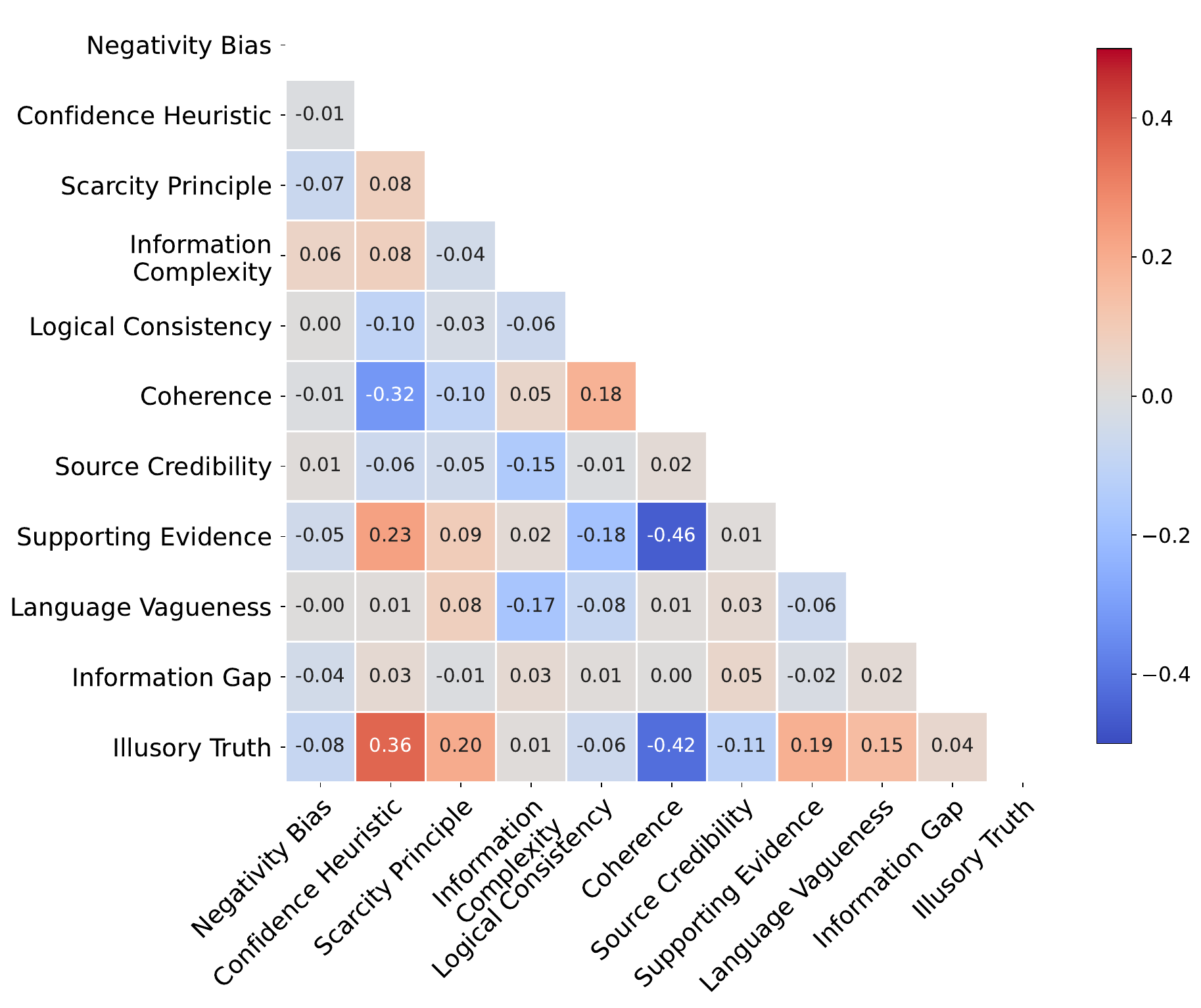}}\vspace{-1em}
  \caption{Feature correlation heatmap on the \textsc{ReCOVery} dataset. Most pairwise correlations fall within a range around zero, indicating that the theory-informed features capture unique information.}
  \label{heatmap}\vspace{-1em}
\end{figure}

To further understand the contribution of each theory to fake news detection, we conducted a feature importance analysis. As shown in Figure \ref{rank} on the \textsc{ReCOVery} dataset, \textit{source credibility} receives high importance when publisher metadata is included. This feature captures the perceived trustworthiness and reputation of news sources, a factor that has been consistently identified across communication, psychology, and information science as the most fundamental predictor of information credibility. However, this result should not be interpreted as evidence that publisher credibility is intrinsically the strongest determinant of misinformation. Publisher identities might be strongly associated with labels in some benchmark datasets. Similarly, \textit{information complexity}, which captures the degree of cognitive demand and structural complexities in the text, ranked second in importance, reinforcing the relevance of information complexity in the context of misinformation. Beyond these top two features, the remaining theory-driven features—including \textit{language vagueness} and \textit{coherence}—show relatively comparable contributions, indicating that multiple theoretical features contribute jointly to the downstream task of fake news detection.

In addition, by looking at Figure~\ref{heatmap} - the Feature Correlation Heatmap - we observe that the absolute values of most pairwise correlations between features fall below 0.1, indicating that the majority of features exhibit independence. Notably, some feature pairs exhibit very small correlations, which reinforces that each feature contributes unique information to some extent. Such a distribution of correlations further demonstrates the validity of incorporating a diverse set of theoretically grounded features, as they provide complementary cues to help detect misinformation. 

\begin{figure}[tbp]
  \centering
  \makebox[\columnwidth][c]{%
  \includegraphics[width=1\columnwidth]{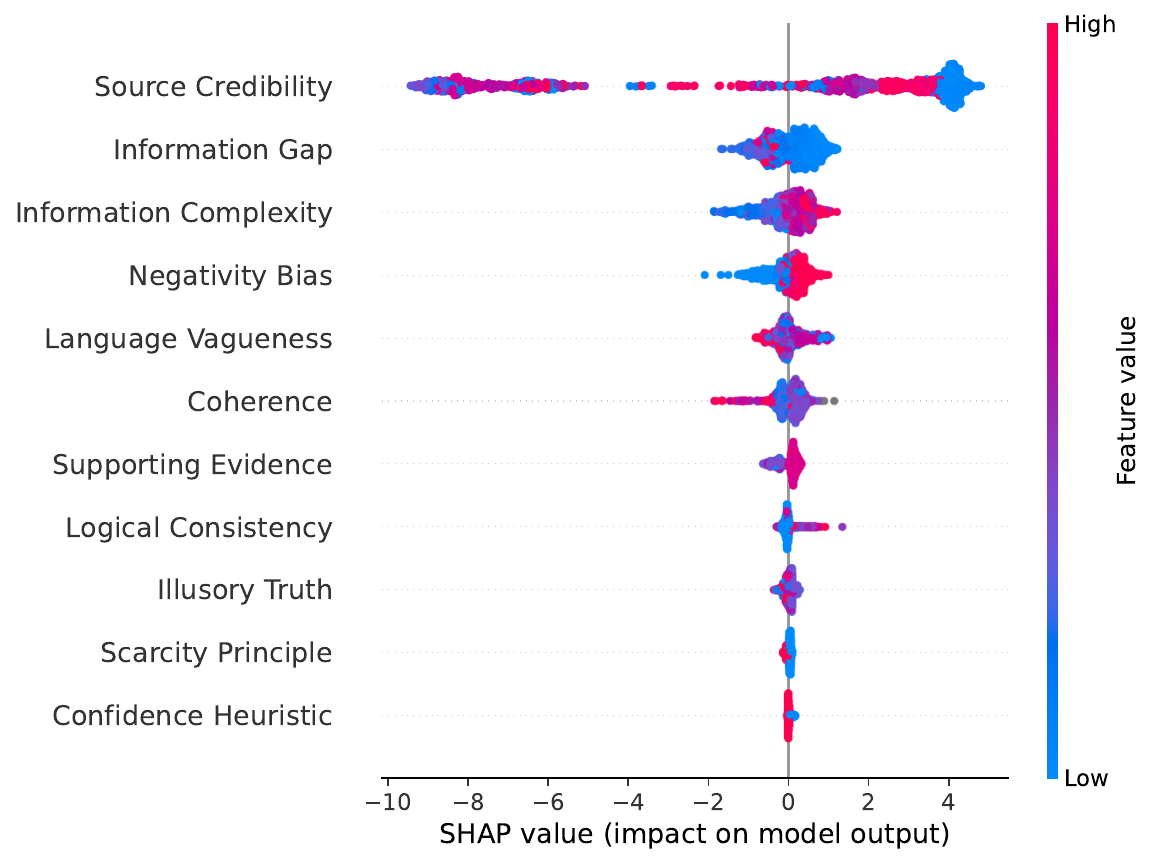}}\vspace{-1em}
    \caption{SHAP plot on the \textsc{ReCOVery} dataset. \textit{Source credibility} stands out as the primary criterion for differentiating fake news from real news.}
  \label{shap}
\end{figure}

\begin{figure*}[htbp]
  \centering
  \includegraphics[width=\linewidth]{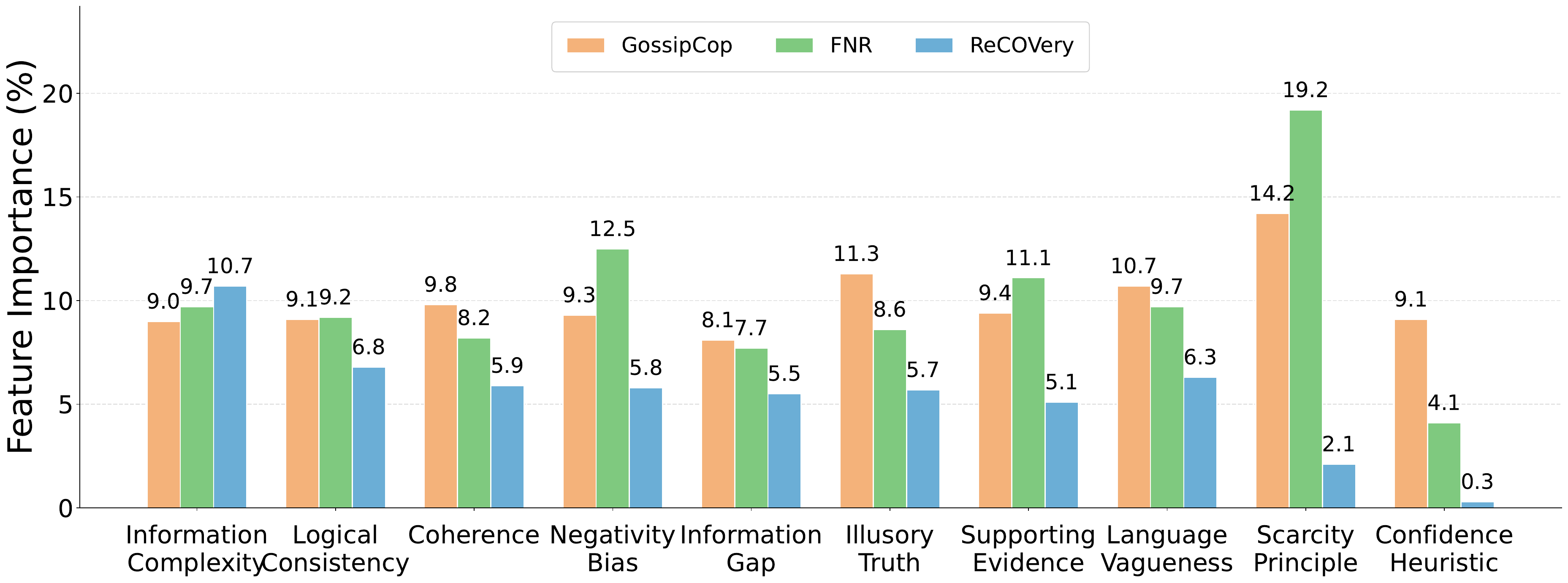}\vspace{-1em}
    \caption{Comparing the importance of theories across datasets.}
  \label{figure:acrossdataset}
\end{figure*}

To further interpret the contribution of each feature, we employ \textit{Shapley Additive exPlanations (SHAP)}, a technique that determines the contribution of each feature to a certain prediction group \cite{lundberg2017unified}. As shown in Figure~\ref{shap}, source credibility consistently emerges as the most decisive factor for distinguishing fake news from real news, and these patterns align with prior research emphasizing the critical role of source credibility in news evaluation. \textit{Information complexity} and \textit{negativity bias} illustrate the most consistent influence on model predictions, with higher values (in red) pushing the prediction toward the misinformation class, whereas lower values (in blue) often shift predictions in the opposite direction. Other theories, such as \textit{information gap} and \textit{coherence}, also demonstrate noticeable but less dominant effects. 
To sum up, these results highlight that a diverse set of theoretically motivated features contributes complementary information to fake news detection. This also shows that SHAP, compared to simple feature ranking, is capable of revealing the directionality of feature impacts on predictions.

\subsection{Cross-dataset comparison}

Figure \ref{figure:acrossdataset} shows feature importance learned by the XGBoost classifier across three datasets. A consistent pattern emerges: source credibility is the single most decisive signal wherever publisher metadata is available (97.2\% on \textsc{GossipCop}; 45.8\% on \textsc{ReCOVery}). That said, we excluded source credibility from our cross-dataset evaluation plot for two reasons. First, this dimension is absent in \textsc{Fake\_And\_Real\_News}—the dataset lacks reliable publisher venue annotations, so the corresponding feature illustrates negligible importance. Second, since source credibility is derived from diverse news websites in \textsc{GossipCop}, this feature dominates performance and obscures the influence of other features. We present in \href{https://docs.google.com/document/d/e/2PACX-1vRMkeNCNq_wgvkoWqxAHcdmDfPObzsxsrLGKopUyRM_-_IozWaJirBe4kzI7sRySZ7lg6Cqzbwc9_od/pub}{\textcolor{blue}{Appendix D}} a detailed analysis of the remaining features.

Beyond source credibility, several content-based theories show stable and moderate contributions across datasets—e.g., information complexity, logical consistency, coherence, and negativity bias—indicating that cognitive and linguistic signals generalize beyond domains. We also observe heterogeneity across datasets: for instance, the scarcity principle is more prominent in \textsc{Fake\_And\_Real\_News}, suggestive of stronger urgency/limited-availability rhetoric in that dataset. Overall, these findings show that several theory-driven features remain predictive across datasets while others vary with specific dataset characteristics such as publisher-level metadata.

\subsection{Potential Patterns of Fake News Content}\label{4.4-pattern}
\begin{figure*}[tbp]
  \centering
  \includegraphics[width=\linewidth]{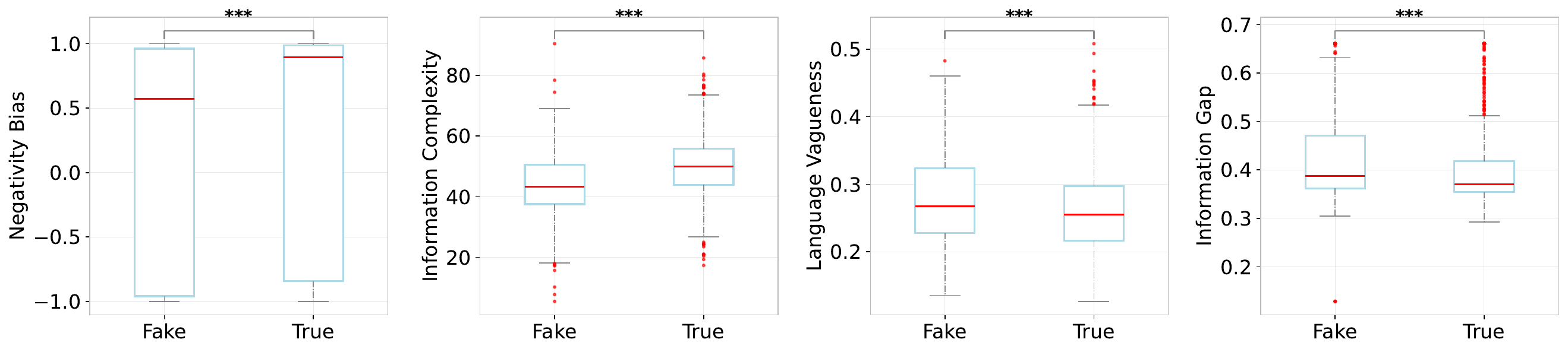}\vspace{-1em}
    \caption{Distributions of four representative theory-informed features across real and fake news.}
  \label{figure:boxplot}
\end{figure*}
To further illustrate the underlying mechanisms differentiating fake news from real news, we examine their content characteristics across several theory-driven dimensions, as shown in Figure \ref{figure:boxplot}. All differences between fake and real news were tested using the \textit{Wilcoxon rank sum test}, with significance levels indicated as``\textbf{***}'' ($p < 0.001$). Specifically, Fake news tends to exhibit significantly lower sentiment scores (reflecting a stronger negativity bias), suggesting more negative framing compared to real news. Fake news also shows lower information complexity, indicating simpler narrative structures and reduced lexical and syntactic diversity. Moreover, fake news illustrates greater linguistic vagueness, revealing the frequent use of ambiguous or imprecise expressions, perhaps to obscure factual details and mislead readers' judgments of truthfulness. Finally, fake news demonstrates a larger information gap, characterized by the use of more uncertainty-inducing cues in headlines—such as question marks, ellipses, exclamation marks, and unexpected or surprising topics—likely designed to arouse curiosity and draw readers' attention. Such patterns empirically reinforce the importance of content-level cues in credibility detection.

\subsection{Cross-Theoretical Interaction Analysis}\label{4.5-crossfeature}

To further investigate how different theoretical dimensions interact to improve fake news detection, we evaluate the predictive performance of single, pairwise, and triple combinations of theory-informed features on the \textsc{ReCOVery}. Single-feature results reveal a clear hierarchy in discriminative power, where \textit{supporting evidence} achieves the highest AUC (0.742), followed by the \textit{confidence heuristic} (0.709). These results underscore the critical role of sourcing cues and persuasive language markers in distinguishing misinformation from credible news stories. The single-feature experiments measure how well each representation predicts the label when used alone, whereas Figure~\ref{rank} measures its contribution in the presence of all other features. Consequently, a representation may be highly predictive in isolation yet receive relatively low importance in the joint model when other features capture overlapping or conditionally substitutable
information.

\begin{figure}[tbp]
 \centering
  \makebox[\columnwidth][c]{%
  \includegraphics[width=1.02\columnwidth]{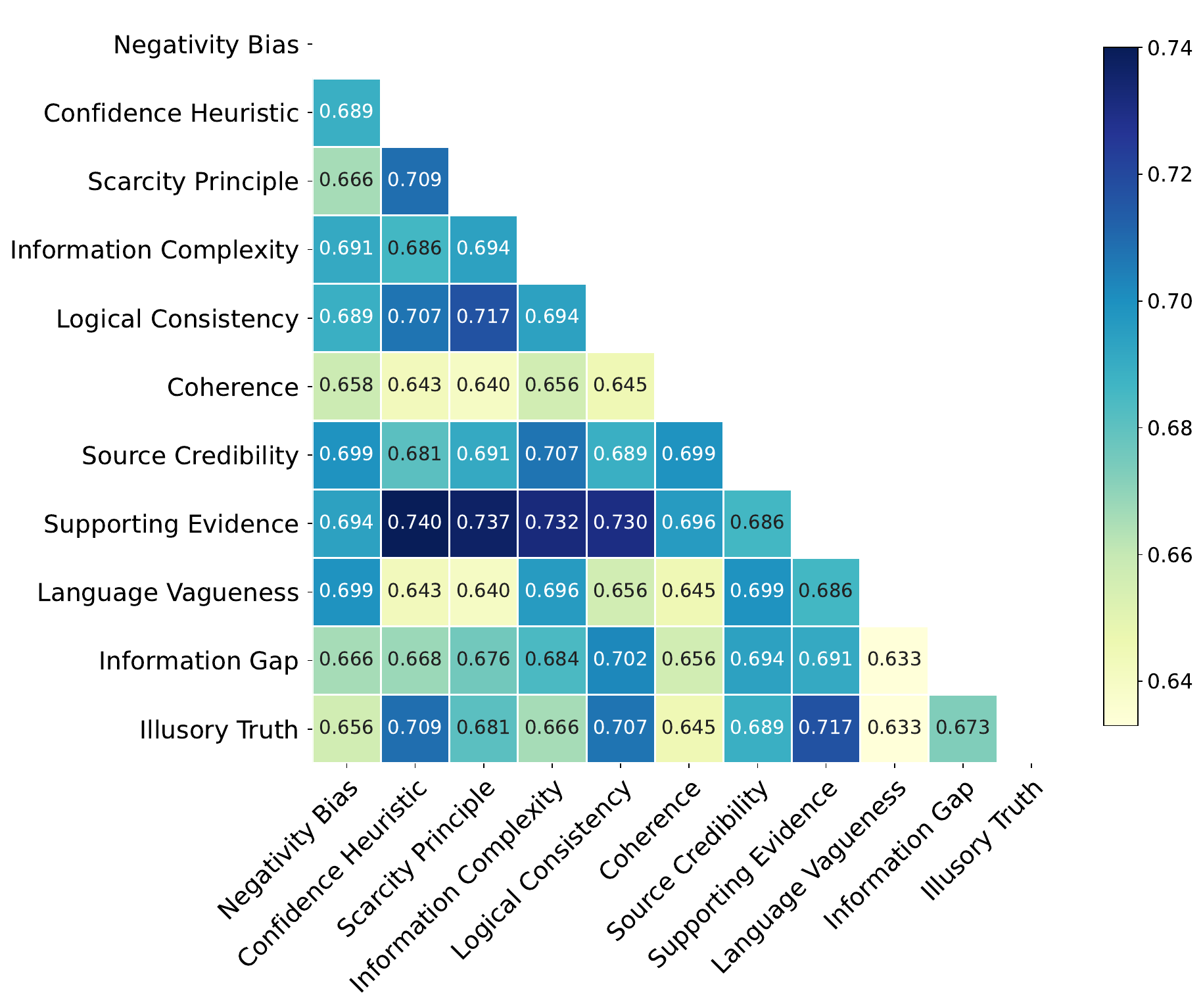}}\vspace{-1em}
    \caption{Pairwise accuracy heatmap illustrating the synergistic effects between theory-informed features.}
  \label{figure:2features}\vspace{-1em}
\end{figure}

The interaction analysis reveals a more complex pattern than a simple synergy account. Particularly, from the pairwise synergy heatmap Figure \ref{figure:2features}, \textit{confidence heuristic} + \textit{supporting evidence} (0.740) and \textit{scarcity principle + supporting evidence} (0.737) represent the most effective two-feature combinations, suggesting that assertive language (reflecting certainty or urgency) and verifiable in-text evidence jointly capture the persuasive intent of fake news content. Similarly, \textit{information complexity + supporting evidence} (0.732) highlights that structural organization and external attribution mutually reinforce credibility cues.

Extending to three-feature configurations, the best combination—(\textit{confidence heuristic}, \textit{scarcity principle}, \textit{supporting evidence})—achieves a classification AUC of 0.745, surpassing any single feature. This result reflects an interaction among language intensity, perceived urgency, and evidence transparency, jointly modeling the persuasive and credibility dimensions of misinformation. However, while multiple theoretical dimensions can provide complementary predictive information, the benefit of combining features is not uniformly additive and should not be interpreted as evidence of strong synergistic effects. Overall, these findings demonstrate that the integration of theory-aligned features produces modest performance improvements and provide empirical evidence that multi-theoretical signals interact non-linearly to enhance model interpretability.

To further assess whether the interaction analysis of theoretical features generalizes across domains, we compare the feature ranking orders between the \textsc{ReCOVery} and \textsc{Fake\_And\_Real\_News} datasets. Specifically, we compute the rank correlations of feature importance derived from the two datasets. The results show a Spearman rank correlation of 0.167 and a Kendall Tau correlation of 0.111, indicating weak alignment between the two feature hierarchies. This discrepancy suggests that while certain theories—such as language intensity—maintain predictive value across datasets, their relative importance is not universal and is affected by dataset-specific factors such as topic domain, publisher diversity, and stylistic conventions. These findings highlight that theory-informed features exhibit dataset-dependent heterogeneity and further underscore the significance of multi-theoretical integration.

\section{Discussion and Limitations}
Our work contributes a structured mapping between social-science theories and computationally measurable signals of news credibility. Theoretically, the framework connects perspectives from persuasion, credibility, language use, and information processing that are often studied separately. Methodologically, it demonstrates how these theories can be transformed into features using lexicon-based measures, statistical machine learning, and large language models, enabling the predictive role and directionality of theory-informed signals rather than being represented only through latent model states. 

A few limitations should be taken into account when interpreting our findings. First, the datasets differ in topic, class distribution, and source composition, which may contribute to dataset-specific feature importance. Then, some representations rely on pretrained or large language models and may be sensitive to prompting and domain shift. Future work could systematically evaluate the robustness of the proposed representations across different prompting strategies and application domains, while also extending the framework to propagation-aware datasets.

\section{Conclusion}
In this work, we develop a theory-informed framework that translates cross-disciplinary perspectives from communication, psychology, economics, and related social sciences into computationally measurable representations for fake-news detection. Rather than relying on high-dimensional latent or lexical representations, our framework connects each feature to a theoretically motivated concept, which could illustrate its predictive contribution and directionality. Our experiments across multiple benchmark datasets demonstrate that theory-derived features contain meaningful diagnostic information for distinguishing fake from real news, although relative importance and effectiveness vary across datasets.  Our goal is not to outperform a full-text representation. Instead, we investigate how much predictive information can be retained using a compact set of theoretically grounded and human-interpretable signals. The resulting framework therefore represents a complementary trade-off between predictive performance and theoretical transparency. As a result, the empirical findings should be interpreted as evidence that theory-informed signals are diagnostic of news veracity labels. The results demonstrate the practical value of connecting social-science theory with computational modeling and provide a transparent basis for analyzing which theoretical cues contribute to automated credibility assessment.

\bibliographystyle{IEEEtran}
\bibliography{IEEEexample}

\end{document}